%% file: main.tex
\documentclass{article}

\usepackage{arxiv}

\usepackage[utf8]{inputenc} 
\usepackage[T1]{fontenc}    
\usepackage{hyperref}       
\usepackage{url}            
\usepackage{booktabs}       
\usepackage{amsfonts}       
\usepackage{nicefrac}       
\usepackage{microtype}      
\usepackage{multirow}
\usepackage{array}
\usepackage{amsmath}
\usepackage{graphicx}
\usepackage[numbers,sort&compress]{natbib}
\usepackage{doi}

\newcommand{\hanORCID}{0000-0003-1834-076X}
\newcommand{\orcid}[2]{%
  \if\relax\detokenize{#1}\relax #2%
  \else\href{https://orcid.org/#1}{\includegraphics[scale=0.06]{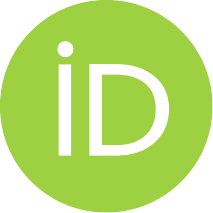}\hspace{1mm}#2}\fi}

\title{Optimal feature selection for firewall log analysis using Machine learning and Hybrid Metaheuristic algorithms}

\author{
	\orcid{\hanORCID}{Seungwoo Han} \\
	Tokyo University of\\
	Agriculture and Technology\\
	Tokyo, Japan \\
	\texttt{seungwoohan0108@gmail.com} \\
	\And
	Gil Hong \\
	Sahmyook University\\
	Seoul, Korea \\
	\texttt{wideroad95@nate.com} \\
	\And
	Jewan Kim \\
	Sahmyook University\\
	Seoul, Korea \\
	\texttt{wpdhks6@gmail.com} \\
	\AND
	Jeuk Yu \\
	Sahmyook University\\
	Seoul, Korea \\
	\texttt{jipsina71@gmail.com} \\
	\And
	Sangjun Lee \\
	Sahmyook University\\
	Seoul, Korea \\
	\texttt{tjdtlf1234@naver.com} \\
	\And
	Byeongok Cho \\
	Sahmyook University\\
	Seoul, Korea \\
	\texttt{winjbo1021@gmail.com} \\
	\AND
	Jusung Jeon \\
	Sahmyook University\\
	Seoul, Korea \\
	\texttt{gggg1371@naver.com} \\
}

\renewcommand{\shorttitle}{Feature selection for firewall log analysis with QBSO}

\hypersetup{
pdftitle={Optimal feature selection for firewall log analysis using Machine learning and Hybrid Metaheuristic algorithms},
pdfsubject={cs.CR, cs.LG},
pdfauthor={Seungwoo Han, Gil Hong, Jewan Kim, Jeuk Yu, Sangjun Lee, Byeongok Cho, Jusung Jeon},
pdfkeywords={bee swarm optimization, reinforcement learning, feature selection, firewall log, data leakage},
}

\begin{document}
\maketitle

\input{ch0_abstract}


\input{ch1_introduction}

\input{ch2_methods}
\input{ch3_results}
\input{ch4_discussion}
\input{ch5_conclusion}

\section*{Code and data availability}
The Internet Firewall Data Set used in this study is publicly available from the UCI Machine Learning Repository at \url{https://archive.ics.uci.edu/dataset/542/internet+firewall+data}~\citep{uci_firewall}. For the implementation of QBSO-FS, please refer to \url{https://github.com/amineremache/qbso-fs}~\citep{remache_qbsofs_code}.

\bibliographystyle{unsrtnat}
\bibliography{references}

\end{document}

%% file: ch0_abstract.tex
\begin{abstract}
Firewall log classification is important to monitor network traffic. Machine learning classifiers using network-related features have shown good results for this task, but high-dimensional features slow down classification. In this paper, we apply feature selection using bee swarm optimization with Q-learning-based local search (QBSO-FS) to the UCI Internet Firewall Data Set, together with random forest (RF), k-nearest neighbor (KNN), and Bernoulli na\"{\i}ve Bayes (Bernoulli NB) classifiers. The first version of this preprint selected and evaluated feature subsets on the same 5-fold cross-validation of the full data set, which contains 12.8\% exact duplicates; the reported scores are therefore optimistically biased. In this extended version, we re-evaluate the method with a leakage-free protocol: the data are split into training, validation, and test sets, features are selected using the validation set only, and accuracy and macro-averaged scores are measured once on the untouched test set, for both the raw data and a deduplicated, preprocessed version. On a single stratified split, QBSO-FS selects 5--7 features. The selected subsets clearly improve Macro-F1 over all features for KNN on both data sets and for RF on the deduplicated data (by 8--19 percentage points). For RF on the raw data and for Bernoulli NB, the differences (0.2--1.5 percentage points) may not hold for a different split. RF with 7 selected features reaches 99.90\% accuracy and 94.88\% Macro-F1 on the deduplicated data. On the same split, this exceeds our re-implementations of the deep neural network configurations of previous studies and the feature subset selected by particle swarm optimization in one of them, while the run time saved on this low-dimensional data set is small. These results suggest that the method can support automatic firewall log analysis, while evaluation over multiple splits remains future work.
\end{abstract}

%% file: ch1_introduction.tex
\section{Introduction}
Firewalls ensure that a computer network operates properly and safely. As the protection of personal information~\cite{castaneda2007effect} becomes increasingly important, so does the protection of the networks that carry it. Firewall logs are key evidence for identifying intrusions, including insider and outsider threats~\cite{singh2010approach}. Beyond traditional rule-based analysis, machine learning and deep learning have been applied to log classification and intrusion prevention~\cite{allagi2019analysis,ertam2018classification,buczak2016survey,le2020analyzing,javaid2016deep,sharma2021optimized}. For a quick response, log analysis and intrusion detection should classify traffic efficiently with as few inputs as possible, yet the features used for classification are often chosen based on the researcher's experience. Feature and parameter selection has therefore been studied using information gain, genetic algorithms, and reinforcement learning~\cite{lei2012feature,janisch2019classification,fan2020autofs,sadeg2019qbso}. In this paper\footnote{This preprint is a revised and extended version of our earlier preprint~\cite{han2021engrxiv} (engrXiv, 2021, \url{https://doi.org/10.31224/osf.io/pm3hy}), whose results are reproduced in Section~\ref{sec:res_original}. It is distinct from the preliminary study of \citet{hong2021study}, published in Korean, which selected four features with a decision tree classifier.}, we search for an optimal feature subset using bee swarm optimization combined with reinforcement learning, namely the Q-learning-based bee swarm optimization for feature selection (QBSO-FS)~\cite{sadeg2019qbso}.

In this extended version, we additionally revisit the evaluation protocol of our original experiment. In the original setting, both the fitness used to search for features and the reported performance were computed by cross-validation on the full data set, and the data set itself contains a substantial number of exactly duplicated records. Both factors can leak information from the evaluation data into model and feature selection and make the reported scores optimistic. The contributions of this extension are as follows:
\begin{itemize}
	\item We identify 8,362 exactly duplicated records (12.8\%) in the Internet Firewall Data Set and build a deduplicated, preprocessed version of the data.
	\item We repeat the QBSO-FS feature selection with the hyper-parameters of the original study, using a leakage-free train/validation/test protocol in which features are selected on a separate validation set and performance is reported only on an untouched test set.
	\item We compare random forest (RF), k-nearest neighbor (KNN), and Bernoulli na\"{\i}ve Bayes (Bernoulli NB) classifiers on both the raw and the deduplicated data, with and without feature selection.
	\item We re-evaluate the configurations of previous studies on the same data set under the same protocol, and compare run times for feature selection, training, and inference.
\end{itemize}

%% file: ch2_methods.tex
\section{Materials and methods}

\subsection{Data set}
We used the Internet Firewall Data Set~\cite{uci_firewall,ertam2018classification,frank2010uci} from the University of California, Irvine (UCI) Machine Learning Repository. It contains 65,532 records with 11 features and one of four labels: allow, deny, drop, or reset-both. Table~\ref{tab:profile} shows the class distribution, and Table~\ref{tab:features} lists the features. NAT stands for network address translation; the NAT source and NAT destination ports are the port numbers after translation by the firewall.

\begin{table}[htbp]
	\centering
	\caption{Class distribution of the raw data set}
	\label{tab:profile}
	\begin{tabular}{lr}
		\toprule
		Label & Records \\
		\midrule
		Allow      & 37,640 \\
		Deny       & 14,987 \\
		Drop       & 12,851 \\
		Reset-both & 54 \\
		\midrule
		Total      & 65,532 \\
		\bottomrule
	\end{tabular}
\end{table}

\begin{table}[htbp]
	\centering
	\caption{Features of the raw data set and of the processed data set (Section~\ref{sec:dedup})}
	\label{tab:features}
	\small
	\begin{tabular}{@{}lll@{}}
		\toprule
		Feature & Description & In the processed data \\
		\midrule
		Source port & Source port of the session & unchanged \\
		Destination port & Destination port of the session & unchanged \\
		NAT source port & Source port after NAT & unchanged \\
		NAT destination port & Destination port after NAT & unchanged \\
		Bytes & Total number of bytes & $\log(1+x)$ \\
		Bytes sent & Number of bytes sent & $\log(1+x)$ \\
		Bytes received & Number of bytes received & $\log(1+x)$ \\
		Packets & Total number of packets & $\log(1+x)$ \\
		Packets sent & Number of packets sent (\texttt{pkts\_sent}) & $\log(1+x)$ \\
		Packets received & Number of packets received (\texttt{pkts\_received}) & $\log(1+x)$ \\
		Elapsed time & Duration of the session in seconds & $\log(1+x)$ \\
		\midrule
		\multicolumn{3}{@{}l}{\emph{Derived features (processed data only)}} \\
		NAT-source-is-zero & 1 if the NAT source port is 0, otherwise 0 & binary \\
		NAT-destination-is-zero & 1 if the NAT destination port is 0, otherwise 0 & binary \\
		Source port range & Range of the source port (well-known, registered, dynamic) & categorical (0, 1, 2) \\
		Destination port range & Range of the destination port (as above) & categorical (0, 1, 2) \\
		Bytes per packet & $\log(1 + \text{bytes}/\text{packets})$ & continuous \\
		\bottomrule
	\end{tabular}
\end{table}

\subsection{Duplicate records and preprocessing}
\label{sec:dedup}
The raw data set contains 8,362 rows (12.8\%) that are exact duplicates of another row, including the label. When such a data set is split into training and evaluation partitions at random, identical records fall on both sides of the split, and a classifier that memorizes the training set is rewarded for recall rather than generalization. Redundant records of this kind are a known source of biased evaluation in network intrusion data sets~\cite{tavallaee2009nslkdd}, and more generally a form of data leakage~\cite{kaufman2012leakage}. To quantify this effect we additionally built a processed version of the data set as follows:
\begin{enumerate}
	\item Exact duplicate rows were removed, leaving 57,170 records (Table~\ref{tab:profile_dedup}).
	\item The heavy-tailed volume and time features (bytes, bytes sent, bytes received, packets, packets sent, packets received, and elapsed time) were transformed with $\log(1+x)$.
	\item Five derived features were added (Table~\ref{tab:features}): the NAT-source-is-zero and NAT-destination-is-zero flags, the source and destination port ranges defined by the Internet Assigned Numbers Authority (IANA)~\cite{rfc6335} (well-known 0--1023, registered 1024--49151, dynamic 49152--65535), and the bytes per packet.
\end{enumerate}
All transformations are applied row by row and do not use statistics estimated from the data, so they do not introduce leakage across partitions. The derived features were defined a priori from domain knowledge of ports and NAT, before any of the experiments reported here, and were not adjusted based on the results. The processed data set therefore has 16 features. Finally, 9 distinct feature vectors occur with more than one label (46 rows in the raw data and 18 rows after deduplication, mostly deny versus drop). These rows are not duplicates and were kept, since removing them would require deciding which label is correct.

\begin{table}[htbp]
	\centering
	\caption{Class distribution before and after deduplication}
	\label{tab:profile_dedup}
	\begin{tabular}{lrrr}
		\toprule
		Label & Raw & Deduplicated & Removed \\
		\midrule
		Allow      & 37,640 & 37,439 & 201 \\
		Deny       & 14,987 & 8,042  & 6,945 \\
		Drop       & 12,851 & 11,635 & 1,216 \\
		Reset-both & 54     & 54     & 0 \\
		\midrule
		Total      & 65,532 & 57,170 & 8,362 \\
		\bottomrule
	\end{tabular}
\end{table}

\subsection{Feature selection with QBSO-FS}

\subsubsection{Bee swarm optimization algorithm}
The bee swarm optimization algorithm (BSO)~\cite{drias2005cooperative} is inspired by the foraging behavior of bees. Each bee explores the search space from its own starting point, and the bees cooperate by sharing the solutions they find over a number of iterations. BSO was later adapted to feature selection as BSO-FS~\cite{sadeg2015bsofs}, and extended with Q-learning-based local search as QBSO-FS~\cite{sadeg2019qbso}, which we use in this work. A solution is a binary vector $\mathbf{s} \in \{0,1\}^n$ that marks the selected features, and its fitness $f(\mathbf{s}) \in [0,1]$ is the accuracy of the classifier trained on those features (Section~\ref{sec:protocol}). The search proceeds as follows; the hyper-parameters named in the steps are listed in Table~\ref{tab:hyper}.
\begin{enumerate}
	\item \emph{Initialization.} A reference solution $\mathbf{s}_{\text{ref}}$ is drawn at random, with each feature selected with probability 0.2.
	\item \emph{Search area.} In each iteration, $\mathbf{s}_{\text{ref}}$ is added to a tabu list $\mathcal{T}$, and $B$ starting solutions, one per bee, are generated from it, where $B$ is the Number of bees. The Flip parameter $F$ controls how far these solutions are from $\mathbf{s}_{\text{ref}}$. Bee $b$ ($b = 0, 1, \dots, F-1$) flips every $F$-th feature starting at position $b$,
	\begin{equation}
		\mathbf{s}^{(b)} = \mathbf{s}_{\text{ref}} \oplus \sum_{m \,:\, mF + b < n} \mathbf{e}_{mF + b},
		\label{eq:searcharea}
	\end{equation}
	where $\oplus$ is the element-wise exclusive OR (XOR) and $\mathbf{e}_j$ is the $j$-th unit vector, so that different bees start in different, non-overlapping directions. If $B > F$, the next $F$ bees each flip a contiguous block of about $n/F$ features, and any remaining bees flip one randomly chosen feature. With the settings of Table~\ref{tab:hyper} ($B = 3$, $F = 5$), all three bees are generated by Eq.~\eqref{eq:searcharea}.
	\item \emph{Local search.} Each bee improves its starting solution by a local search. In QBSO-FS this is the Q-learning procedure of Section~\ref{sec:rl}, whose number of steps is governed by Local iteration.
	\item \emph{Communication.} The bees report their solutions and fitness values to a shared ``dance table'', from which the next reference solution is chosen. If the best bee improves on the best solution found so far, or on the last solution in $\mathcal{T}$, it becomes the new $\mathbf{s}_{\text{ref}}$ and the number of remaining chances is reset to Max chance. Otherwise, one chance is consumed. While chances remain, the best bee is still chosen (intensification). When they are exhausted, the bee farthest in Hamming distance from the solutions in $\mathcal{T}$ is chosen instead (diversification), which lets the swarm escape a local optimum. Ties in fitness are broken in favor of the solution farther from $\mathcal{T}$ and with fewer features.
	\item \emph{Termination.} Steps 2--4 are repeated for Max iteration iterations, and the best solution found is returned.
\end{enumerate}
The total number of local searches is therefore $B \times$ Max iteration, and the search budget is controlled mainly by these two parameters together with Local iteration.

\subsubsection{Q-learning local search}
\label{sec:rl}
Reinforcement learning (RL) trains an agent to choose actions that maximize its cumulative reward. QBSO-FS uses Q-learning~\cite{watkins1992qlearning} as the local search of each bee. The state is the current feature subset, an action flips one feature, and the reward favors higher accuracy and, at equal accuracy, fewer features. To reduce the number of candidate actions, only features on which the current state and the best solution differ are considered. We now state the procedure formally.

\paragraph{Formulation.}\label{sec:rl_formulation}
Let $n$ be the number of features. A state is a binary vector
\begin{equation}
	\mathbf{s} = (s_1, \dots, s_n) \in \{0,1\}^n, \qquad |\mathbf{s}| = \sum_{j=1}^{n} s_j ,
\end{equation}
where $s_j = 1$ means that feature $j$ is selected. An action $a \in \{1, \dots, n\}$ flips one feature:
\begin{equation}
	\delta(\mathbf{s}, a) =
	\begin{cases}
		\mathbf{s} \oplus \mathbf{e}_a & \text{if } |\mathbf{s} \oplus \mathbf{e}_a| > 0, \\
		\mathbf{s} & \text{otherwise},
	\end{cases}
\end{equation}
so the empty subset is never visited.

\emph{Action space reduction.} Let $\mathbf{s}^{*}$ be the best solution found so far by the swarm. Only features on which the current state and the best solution disagree, as determined by their XOR, are considered:
\begin{equation}
	D(\mathbf{s}) = \{\, j \mid s_j \oplus s^{*}_j = 1 \,\}.
\end{equation}
The candidate action set $\mathcal{A}(\mathbf{s})$ is a uniformly random subset of $D(\mathbf{s})$ of size $F$ if $|D(\mathbf{s})| > F$, and a single random element of $D(\mathbf{s})$ otherwise. If $D(\mathbf{s})$ is empty, the previous action set is kept.

\emph{Policy.} Actions are chosen $\epsilon$-greedily with respect to the Q-value plus the fitness of the resulting state:
\begin{equation}
	a_t =
	\begin{cases}
		\displaystyle \arg\max_{a \in \mathcal{A}(\mathbf{s}_t)} \bigl[\, Q(\mathbf{s}_t, a) + f\bigl(\delta(\mathbf{s}_t, a)\bigr) \bigr] & \text{with probability } 1-\epsilon, \\
		\text{uniform random } a \in \mathcal{A}(\mathbf{s}_t) & \text{with probability } \epsilon,
	\end{cases}
\end{equation}
with ties broken at random. After each step, the exploration rate decays linearly, $\epsilon \leftarrow \max(0,\ \epsilon - 10^{-4})$.

\emph{Reward.} With $\mathbf{s}_{t+1} = \delta(\mathbf{s}_t, a_t)$, the reward is
\begin{equation}
	r_t =
	\begin{cases}
		f(\mathbf{s}_{t+1}) & \text{if } f(\mathbf{s}_{t+1}) > f(\mathbf{s}_t), \\
		f(\mathbf{s}_{t+1}) - f(\mathbf{s}_t) & \text{if } f(\mathbf{s}_{t+1}) < f(\mathbf{s}_t), \\
		\tfrac{1}{4}\, f(\mathbf{s}_{t+1}) & \text{if } f(\mathbf{s}_{t+1}) = f(\mathbf{s}_t) \text{ and } |\mathbf{s}_{t+1}| < |\mathbf{s}_t|, \\
		-\tfrac{1}{4}\, f(\mathbf{s}_{t+1}) & \text{if } f(\mathbf{s}_{t+1}) = f(\mathbf{s}_t) \text{ and } |\mathbf{s}_{t+1}| \ge |\mathbf{s}_t|.
	\end{cases}
	\label{eq:reward}
\end{equation}
The feature-count terms therefore act as a tie-breaker that favors smaller subsets when accuracy does not change.

\emph{Q-value update.} The Q-table is updated with the standard Q-learning rule~\cite{watkins1992qlearning}, using the learning rate $\alpha$ and the discount factor $\gamma$ of Table~\ref{tab:hyper}:
\begin{equation}
	Q(\mathbf{s}_t, a_t) \leftarrow (1-\alpha)\, Q(\mathbf{s}_t, a_t) + \alpha \Bigl[\, r_t + \gamma\, Q(\mathbf{s}_{t+1}, a') \Bigr],
	\qquad
	a' = \arg\max_{a \in \mathcal{A}} \bigl[\, Q(\mathbf{s}_{t+1}, a) + f\bigl(\delta(\mathbf{s}_{t+1}, a)\bigr) \bigr],
	\label{eq:qupdate}
\end{equation}
where $a'$ is the greedy action in the next state under the same criterion as the policy. All Q-values are initialized to zero, and the Q-table is shared by all bees, so the experience gathered by one bee guides the local search of the others.

\emph{Local search schedule.} In BSO iteration $i$, each bee performs
\begin{equation}
	K_i = \min\bigl(\lfloor i / L \rfloor + 1,\ L\bigr)
\end{equation}
Q-learning steps, where $L$ is the Local iteration parameter. With $L = 2$ and three iterations, this gives $K = 1, 2, 2$, so the search explores in the first iteration and exploits more in later ones. The fitness of a bee is $f(\mathbf{s})$ of its final state.

\subsubsection{Hyper-parameters}
Table~\ref{tab:hyper} lists the hyper-parameters of QBSO-FS, which are the same in the original and in the leakage-free experiment.

\begin{table}[htbp]
	\centering
	\caption{Hyper-parameters of QBSO-FS}
	\label{tab:hyper}
	\begin{tabular}{llp{0.55\linewidth}}
		\toprule
		Name & Value & Description \\
		\midrule
		Flip ($F$)            & 5    & Spacing of the flipped features in the search area (Eq.~\eqref{eq:searcharea}) and size of the candidate action set \\
		Max chance            & 3    & Iterations without improvement allowed before diversification \\
		Number of bees ($B$)  & 3    & Number of bees, i.e.\ starting solutions per iteration \\
		Max iteration         & 3    & Number of BSO iterations \\
		Local iteration ($L$) & 2    & Upper bound on the Q-learning steps per bee and iteration \\
		$\alpha$              & 0.8  & Learning rate of Q-learning \\
		$\gamma$              & 0.3  & Discount factor of Q-learning \\
		$\epsilon$            & 0.05 & Initial probability of a random action \\
		\bottomrule
	\end{tabular}
\end{table}

\subsubsection{Why BSO for wrapper feature selection}
Wrapper methods evaluate each candidate subset by training the classifier itself~\cite{kohavi1997wrappers}, so the search strategy determines both the quality of the result and the number of costly evaluations. BSO-FS has also been applied in network security, for example to select features for detecting external sinkhole attacks in large-scale wireless sensor networks~\cite{senior2026svm}. Compared with common alternatives, BSO, and QBSO-FS in particular, has the following properties.
\begin{itemize}
	\item \emph{Escaping local optima.} Greedy sequential methods such as forward or backward selection never revisit a decision and can be trapped in a local optimum (the nesting effect)~\cite{pudil1994floating}. BSO keeps a tabu list~\cite{glover1989tabu} of past reference solutions and switches from intensification to diversification once Max chance iterations pass without improvement. It therefore explicitly moves away from regions that have already been explored.
	\item \emph{Native binary search.} A feature subset is a bit vector, and BSO operates on it directly through bit flips (Eq.~\eqref{eq:searcharea}). Particle swarm optimization (PSO)~\cite{kennedy1995pso} was designed for continuous spaces and needs a transfer function that maps velocities to bit probabilities in its binary version~\cite{kennedy1997binarypso}. Genetic algorithms (GA)~\cite{siedlecki1989ga} also work on bit strings, but their crossover and mutation operators are random.
	\item \emph{Systematic, diverse neighborhoods.} The Flip-based search area assigns each bee a different, non-overlapping set of features to change. A few bees thus cover distinct directions around the reference solution, and the population can be very small (three bees in this work). With a population-based method such as GA or PSO, evaluating a whole population in every generation is the dominant cost of a wrapper.
	\item \emph{Learning from experience.} The local search is guided by a Q-table shared by all bees (Eq.~\eqref{eq:qupdate}), so flips that improved the fitness before are preferred later. The XOR-based action space restricts the moves to features on which the current state and the best solution so far disagree.
	\item \emph{Preference for compact subsets.} At equal fitness, the reward (Eq.~\eqref{eq:reward}) and the tie-breaking rule of the dance table both favor solutions with fewer features. Wrappers that maximize accuracy alone do not include this preference.
\end{itemize}
These are structural advantages and do not guarantee better results on every problem~\cite{wolpert1997nfl}. In this study we did not run GA or PSO under the same evaluation budget. The empirical evidence here is limited to the comparison with all features and with the subset reported for tuned particle swarm optimization (TPSO)~\cite{verma2024tpso} (Section~\ref{sec:res_prior}). In our experiments, QBSO-FS needed 270--551 subset evaluations (Table~\ref{tab:runtime}), a small fraction of the $2^{11}-1 = 2{,}047$ and $2^{16}-1 = 65{,}535$ non-empty subsets of the raw and processed data.

\subsection{Classifiers}
We compare three classifiers: RF~\cite{breiman2001rf}, KNN~\cite{cover1967knn}, and Bernoulli NB~\cite{mccallum1998nb}. Bernoulli NB binarizes every input at a threshold of 0, so each feature is reduced to a zero/non-zero indicator. Default hyper-parameters are used for all classifiers.

\subsection{Evaluation protocol}
\label{sec:protocol}
\paragraph{Evaluation metrics.}
Let $h$ be a trained classifier and $\mathcal{S}$ a set of $N$ labeled records, with true labels $y_i$ and predicted labels $\hat{y}_i$, and let $\mathcal{C}$ be the set of $C = 4$ classes. For each class $c \in \mathcal{C}$, $\mathrm{TP}_c$, $\mathrm{FP}_c$, and $\mathrm{FN}_c$ denote the numbers of true positives, false positives, and false negatives when $c$ is treated as the positive class. Accuracy (Acc.) is the fraction of correctly classified records,
\begin{equation}
	\operatorname{Acc}(h, \mathcal{S}) = \frac{1}{N} \sum_{i=1}^{N} \mathbf{1}\left[\hat{y}_i = y_i\right] = \frac{1}{N} \sum_{c \in \mathcal{C}} \mathrm{TP}_c ,
\end{equation}
and the per-class precision (P), recall (R), and F1 score are
\begin{equation}
	P_c = \frac{\mathrm{TP}_c}{\mathrm{TP}_c + \mathrm{FP}_c}, \qquad
	R_c = \frac{\mathrm{TP}_c}{\mathrm{TP}_c + \mathrm{FN}_c}, \qquad
	\mathrm{F1}_c = \frac{2\, P_c\, R_c}{P_c + R_c}.
\end{equation}
The macro-averaged precision (Macro-P), recall (Macro-R), and F1 score (Macro-F1) give each class the same weight regardless of its size:
\begin{equation}
	\text{Macro-P} = \frac{1}{C} \sum_{c \in \mathcal{C}} P_c, \qquad
	\text{Macro-R} = \frac{1}{C} \sum_{c \in \mathcal{C}} R_c, \qquad
	\text{Macro-F1} = \frac{1}{C} \sum_{c \in \mathcal{C}} \mathrm{F1}_c .
\end{equation}
Macro-F1 is thus the mean of the per-class F1 scores, not the F1 score of Macro-P and Macro-R. If a class is never predicted, its precision (and hence its F1 score) is set to zero. For previous studies whose averaging is not stated (Table~\ref{tab:prior_reported}), P, R, and F1 refer to the values reported by the authors. Because the classes are highly imbalanced, Acc.\ is dominated by the allow class, whereas the macro-averaged scores are sensitive to the 54-record reset-both class. Differences between scores are given in percentage points (pp).

\paragraph{Original protocol.} In our original experiment, the fitness of a candidate feature subset was the mean accuracy of stratified 5-fold cross-validation (CV) over the \emph{entire} data set, and the performance reported in Section~\ref{sec:res_original} was obtained from the same folds. Because the subset is chosen to maximize exactly the score that is then reported, this estimate is optimistically biased (selection bias~\cite{ambroise2002selection}), and the duplicated records described in Section~\ref{sec:dedup} further inflate it. Formally, with $\mathcal{D}$ the full data set split into folds $\mathcal{D}_1,\dots,\mathcal{D}_5$,
\begin{equation}
	f_{\text{orig}}(\mathbf{s}) = \frac{1}{5} \sum_{k=1}^{5} \operatorname{Acc}\bigl(h_{\mathbf{s}}^{(\mathcal{D} \setminus \mathcal{D}_k)},\ \mathcal{D}_k\bigr),
\end{equation}
where $h_{\mathbf{s}}^{(\mathcal{S})}$ denotes the classifier trained on set $\mathcal{S}$ using only the features selected by $\mathbf{s}$.

\paragraph{Leakage-free protocol.} For the additional experiment we use a held-out design. Each data set is split, stratified by label, into a training set (60\%), a validation set (20\%), and a test set (20\%). During feature selection, the fitness of a subset is the accuracy on the validation set of a classifier fitted on the training set,
\begin{equation}
	f_{\text{val}}(\mathbf{s}) = \operatorname{Acc}\bigl(h_{\mathbf{s}}^{(\mathcal{D}_{\text{train}})},\ \mathcal{D}_{\text{val}}\bigr),
\end{equation}
and the test set is not touched during the search. After the search, the classifier is refitted on the union of the training and validation sets using the selected features and evaluated once on the test set, i.e.\ we report $\operatorname{Acc}\bigl(h_{\hat{\mathbf{s}}}^{(\mathcal{D}_{\text{train}} \cup \mathcal{D}_{\text{val}})},\ \mathcal{D}_{\text{test}}\bigr)$ and the macro-averaged scores for the selected subset $\hat{\mathbf{s}}$. The same refit-and-test procedure with all features serves as the baseline. The protocol is applied to both the raw data (duplicates kept) and the processed, deduplicated data.

\subsection{Implementation}
All classifiers in the leakage-free experiment, including the multilayer perceptrons (MLPs) in Section~\ref{sec:res_prior}, were implemented with scikit-learn~\cite{pedregosa2011sklearn} (version 1.9.1); the original experiment (Section~\ref{sec:res_original}) used version 0.24.2. For the leakage-free experiment, we used the open-source QBSO-FS implementation~\cite{remache_qbsofs_code} of \citet{sadeg2019qbso} with Q-learning local search, the hyper-parameters of Table~\ref{tab:hyper}, and the reward of Eq.~\eqref{eq:reward}.

%% file: ch3_results.tex
\section{Results}

\subsection{Original experiment}
\label{sec:res_original}
This subsection reproduces the results reported in the first version of this preprint~\cite{han2021engrxiv}. They were obtained with the original protocol (Section~\ref{sec:protocol}): feature subsets were selected from the 11 features of the raw data set and scored with the same 5-fold CV on the full, non-deduplicated data, and Table~\ref{tab:bestacc} reports the best score among the candidate subsets. These values are therefore optimistically biased. We keep them for reference only; the leakage-free estimates in Section~\ref{sec:res_holdout} supersede them.

The feature combinations selected from the 11 raw features are shown in Table~\ref{tab:optimal}.

\begin{table}[htbp]
	\centering
	\caption{Feature combinations selected in the original experiment (raw data set, 11 features)}
	\label{tab:optimal}
	\begin{tabular}{lp{0.6\linewidth}}
		\toprule
		Classifier & Selected feature combination \\
		\midrule
		RF & Destination port, NAT source port, NAT destination port, bytes, bytes received, packets, elapsed time, packets sent, packets received \\
		KNN & Destination port, NAT destination port, bytes, bytes sent, packets, elapsed time \\
		Bernoulli NB & Source port, bytes, bytes sent, elapsed time, packets received \\
		\bottomrule
	\end{tabular}
\end{table}

Table~\ref{tab:bestacc} shows, for each classifier, the best accuracy among the candidate subsets.

\begin{table}[htbp]
	\centering
	\caption{Best accuracy (\%) among candidate subsets (original protocol; optimistically biased)}
	\label{tab:bestacc}
	\begin{tabular}{lr}
		\toprule
		Classifier & Best Acc. \\
		\midrule
		RF           & \textbf{99.87} \\
		KNN          & 99.78 \\
		Bernoulli NB & 80.22 \\
		\bottomrule
	\end{tabular}
\end{table}

We also compared RF with the selected features and with all features in terms of Acc., Macro-P, Macro-R, and Macro-F1, each averaged over the five folds. The results are shown in Table~\ref{tab:rf_compare}.

\begin{table}[htbp]
	\centering
	\caption{Comparing all and selected features using RF (\%) (original protocol, 5-fold CV on the full raw data set with 11 features; optimistically biased)}
	\label{tab:rf_compare}
	\begin{tabular}{lrrrr}
		\toprule
		Feature selected & Acc. & Macro-P & Macro-R & Macro-F1 \\
		\midrule
		All     & 99.80 & 93.06 & 81.71 & 84.22 \\
		Selected & \textbf{99.87} & \textbf{96.62} & \textbf{88.73} & \textbf{91.27} \\
		\bottomrule
	\end{tabular}
\end{table}

\subsection{Leakage-free experiment}
\label{sec:res_holdout}
Table~\ref{tab:split} summarizes the partitions used in the leakage-free protocol (Section~\ref{sec:protocol}). For the raw data, 2,365 of the 13,107 test records (18.0\%) have an identical copy (features and label) in the training or validation set. After deduplication, no test record appears in the training or validation set.

\begin{table}[htbp]
	\centering
	\caption{Train / validation / test partitions (stratified 60/20/20)}
	\label{tab:split}
	\begin{tabular}{lrrrrr}
		\toprule
		Data set & Features & Train & Validation & Test & Test rows duplicated in train+val \\
		\midrule
		Raw (duplicates kept) & 11 & 39,318 & 13,107 & 13,107 & 2,365 (18.0\%) \\
		Processed (deduplicated) & 16 & 34,302 & 11,434 & 11,434 & 0 (0.0\%) \\
		\bottomrule
	\end{tabular}
\end{table}

The feature subsets selected on the validation set are listed in Table~\ref{tab:holdout_features}. On the raw data, the subsets chosen for KNN and Bernoulli NB are very close to those of the original experiment (Table~\ref{tab:optimal}): the Bernoulli NB subset is identical, and the KNN subset differs only by replacing packets with bytes received. The RF subset is smaller (5 instead of 9 features). On the processed data, every classifier selects 7 of the 16 features, and each subset contains at least one of the derived features.

\begin{table}[htbp]
	\centering
	\caption{Feature subsets selected on the validation set (leakage-free experiment)}
	\label{tab:holdout_features}
	\begin{tabular}{llcp{0.55\linewidth}}
		\toprule
		Data set & Classifier & \# & Selected features \\
		\midrule
		\multirow{3}{*}{Raw}
		& RF & 5 & Destination port, NAT source port, bytes, bytes received, elapsed time \\
		& KNN & 6 & Destination port, NAT destination port, bytes, bytes sent, bytes received, elapsed time \\
		& Bernoulli NB & 5 & Source port, bytes, bytes sent, elapsed time, packets received \\
		\midrule
		\multirow{3}{*}{Processed}
		& RF & 7 & Destination port, bytes received, packets, elapsed time, packets sent, NAT-destination-is-zero, destination port range \\
		& KNN & 7 & Bytes sent, packets, elapsed time, NAT-source-is-zero, NAT-destination-is-zero, destination port range, bytes per packet \\
		& Bernoulli NB & 7 & Source port, bytes, elapsed time, packets received, source port range, destination port range, bytes per packet \\
		\bottomrule
	\end{tabular}
\end{table}

Table~\ref{tab:holdout_results} reports the test-set performance of each classifier with all features and with the selected subset. ``Val.\ acc.'' is the validation accuracy of the classifier trained on the training set; for the selected subset, it is the fitness maximized by QBSO-FS.

\begin{table}[htbp]
	\centering
	\caption{Test-set performance (\%) of the leakage-free experiment. Classifiers are refitted on train+validation and evaluated once on the held-out test set.}
	\label{tab:holdout_results}
	\begin{tabular}{lllrrrrrr}
		\toprule
		Data set & Classifier & Features & \# & Val.\ acc. & Accuracy & Macro-P & Macro-R & Macro-F1 \\
		\midrule
		\multirow{6}{*}{Raw}
		& \multirow{2}{*}{RF} & All & 11 & 99.82 & 99.78 & \textbf{99.75} & 83.91 & 88.12 \\
		& & Selected & 5 & 99.86 & \textbf{99.84} & 95.65 & \textbf{86.26} & \textbf{89.57} \\
		\cmidrule(l){2-9}
		& \multirow{2}{*}{KNN} & All & 11 & 99.34 & 99.37 & 74.43 & 74.60 & 74.51 \\
		& & Selected & 6 & 99.73 & \textbf{99.76} & \textbf{96.66} & \textbf{90.72} & \textbf{93.22} \\
		\cmidrule(l){2-9}
		& \multirow{2}{*}{Bernoulli NB} & All & 11 & 79.90 & 79.80 & 38.28 & 49.77 & 42.25 \\
		& & Selected & 5 & 80.27 & \textbf{80.19} & \textbf{38.39} & \textbf{49.94} & \textbf{42.42} \\
		\midrule
		\multirow{6}{*}{Processed}
		& \multirow{2}{*}{RF} & All & 16 & 99.75 & 99.79 & 91.35 & 79.35 & 81.89 \\
		& & Selected & 7 & 99.87 & \textbf{99.90} & \textbf{97.11} & \textbf{93.06} & \textbf{94.88} \\
		\cmidrule(l){2-9}
		& \multirow{2}{*}{KNN} & All & 16 & 99.19 & 99.20 & 73.97 & 74.33 & 74.15 \\
		& & Selected & 7 & 99.85 & \textbf{99.83} & \textbf{99.74} & \textbf{79.40} & \textbf{82.49} \\
		\cmidrule(l){2-9}
		& \multirow{2}{*}{Bernoulli NB} & All & 16 & 98.27 & 98.15 & 73.23 & 72.94 & 73.04 \\
		& & Selected & 7 & 99.03 & \textbf{98.91} & \textbf{73.71} & \textbf{73.23} & \textbf{73.43} \\
		\bottomrule
	\end{tabular}
\end{table}

In all six combinations of data set and classifier, the selected subset uses fewer than half of the features, and on this split its test accuracy and Macro-F1 are higher than those of the full feature set. The size of the difference, however, varies widely. It is large for KNN (Macro-F1 $+18.71$\,pp on the raw data and $+8.34$\,pp on the processed data) and for RF on the processed data ($+12.99$\,pp). It is small for RF on the raw data ($+1.45$\,pp, with lower Macro-P than all features) and for Bernoulli NB ($+0.17$\,pp on the raw data and $+0.39$\,pp on the processed data). Differences of this size could be reversed by a different split, so for these cases we can only conclude that the selected subset performs about as well as all features with fewer than half of them. The test accuracy of every selected subset is within 0.12\,pp of its validation fitness, so there is no sign that the search overfitted the validation set. The best result is obtained by RF on the deduplicated data with 7 features (99.90\% accuracy, 94.88\% Macro-F1). Because all numbers come from a single split, differences of a few tenths of a pp in accuracy, and of a few pp in the macro-averaged scores (Section~\ref{sec:discussion}), should not be over-interpreted.

\subsection{Comparison with previous studies}
\label{sec:res_prior}
Table~\ref{tab:prior_reported} summarizes studies that classify the same UCI Internet Firewall Data Set, as reported by their authors. \citet{ertam2018classification}, who collected the data set, classified it with multiclass support vector machines (SVMs) using linear, polynomial, radial basis function (RBF), and sigmoid kernels. \citet{lillmond2021dnn} trained a deep neural network (DNN), implemented as an MLP, on all 11 features. \citet{verma2024tpso} selected 5 features with a TPSO wrapper that uses a Na\"{\i}ve Bayes classifier on the raw feature values as the objective, and then trained a DNN on them. \citet{talaeepoor2024dt} compared a decision tree, KNN, and neural networks under several $k$-fold settings and found the decision tree to perform best. The reported figures are not directly comparable. The studies differ in the number of classes (Lillmond and Suddul merge reset-both into deny and undersample the majority classes), in how features are selected and evaluated (\citet{lillmond2021dnn}, \citet{verma2024tpso}, and \citet{talaeepoor2024dt}, like our original experiment, use CV on the full, non-deduplicated data, and \citet{ertam2018classification} do not report their split), and in how precision, recall, and F1 are averaged.

\begin{table}[htbp]
	\centering
	\caption{Previous studies on the UCI Internet Firewall Data Set, as reported by the authors (\%).}
	\label{tab:prior_reported}
	\small
	\begin{tabular}{@{}>{\raggedright\arraybackslash}p{0.14\linewidth}>{\raggedright\arraybackslash}p{0.16\linewidth}>{\raggedright\arraybackslash}p{0.17\linewidth}>{\raggedright\arraybackslash}p{0.14\linewidth}>{\raggedright\arraybackslash}p{0.13\linewidth}>{\raggedright\arraybackslash}p{0.13\linewidth}@{}}
		\toprule
		Study & Data & Feature selection & Evaluation & Reported result & Reported time \\
		\midrule
		\citet{ertam2018classification} & 4 classes, duplicates kept & None (11 features) & Multiclass SVM (linear, polynomial, RBF, sigmoid); split not reported & Best F1 76.4 (RBF; P 63.0, R 97.1); accuracy not reported & -- \\
		\citet{lillmond2021dnn} & 3 classes (reset-both merged into deny), undersampled & None (11 features) & 5-fold CV, MLP (10, 5) & Acc.\ 94.49, F1 94.97 & -- \\
		\citet{verma2024tpso} & 4 classes, duplicates kept & TPSO wrapper (Na\"{\i}ve Bayes, 10-fold CV on full data), 5 features & 10-fold CV, DNN & Acc.\ 98.29, P 100, R 99.99, F1 99.99 & Model build 845.13\,s (all) / 668.66\,s (TPSO) \\
		\citet{talaeepoor2024dt} & 4 classes, duplicates kept & None (11 features); parameters tuned by grid search & 2-, 5-, 10-, 20-fold CV, decision tree & Acc.\ 99.85, P 94.08, R 87.69, F1 90.77 (2-fold) & -- \\
		Ours, original~\citep{han2021engrxiv} (Section~\ref{sec:res_original}) & 4 classes, duplicates kept & QBSO-FS (5-fold CV on full data), 9 features & 5-fold CV, RF & Acc.\ 99.87, Macro-F1 91.27 & -- \\
		Ours, leakage-free (Section~\ref{sec:res_holdout}) & 4 classes, deduplicated & QBSO-FS (validation set), 7 features & Held-out test, RF & Acc.\ 99.90, Macro-F1 94.88 & Search 96.3\,s, training 0.17\,s \\
		\bottomrule
	\end{tabular}
\end{table}

To compare the approaches on equal terms, we re-ran their configurations under our leakage-free protocol, with the same split and the same four classes (Table~\ref{tab:prior_eval}). We evaluated three feature sets with each model: all features, the 5-feature TPSO subset reported by Verma and Sahu, and the QBSO-FS subset. In addition to RF, KNN, and Bernoulli NB, we included the two DNN configurations of the prior studies, implemented as MLPs with standard scaling. DNN-L follows \citet{lillmond2021dnn} (two hidden layers of 10 and 5 neurons with an L2 penalty), and DNN-V follows \citet{verma2024tpso} (hidden layers of 11 neurons trained with stochastic gradient descent (SGD); because the number of hidden layers is not stated, we used two). We did not run QBSO-FS with a DNN in the loop because of its cost, and the RF subset was reused for the DNNs instead. Table~\ref{tab:runtime} reports the corresponding run times.

\begin{table}[htbp]
	\centering
	\caption{Re-evaluation of prior configurations under the leakage-free protocol (test set, \%). ``TPSO'' denotes the 5-feature subset selected in the TPSO study; ``QBSO'' denotes the 5--7-feature subset selected by QBSO-FS on the validation set for the same classifier (Table~\ref{tab:holdout_features}; for the two DNNs, the RF subset is reused). Best Macro-F1 per row in bold.}
	\label{tab:prior_eval}
	\small
	\begin{tabular}{llrrrrrr}
		\toprule
		& & \multicolumn{2}{c}{All} & \multicolumn{2}{c}{\shortstack{TPSO\\{\scriptsize\citep{verma2024tpso}}}} & \multicolumn{2}{c}{\shortstack{QBSO\\{\scriptsize\citep{sadeg2019qbso}}}} \\
		\cmidrule(lr){3-4}\cmidrule(lr){5-6}\cmidrule(lr){7-8}
		Data set & Model & Acc. & Macro-F1 & Acc. & Macro-F1 & Acc. & Macro-F1 \\
		\midrule
		\multirow{5}{*}{Raw} & RF & 99.78 & 88.12 & 99.69 & 88.63 & 99.84 & \textbf{89.57} \\
		 & KNN & 99.37 & 74.51 & 99.12 & 74.32 & 99.76 & \textbf{93.22} \\
		 & Bernoulli NB & 79.80 & 42.25 & 80.13 & 42.39 & 80.19 & \textbf{42.42} \\
		 & DNN-L (MLP 10--5) & 99.64 & \textbf{74.72} & 99.65 & \textbf{74.72} & 98.12 & 72.93 \\
		 & DNN-V (MLP 11--11, SGD) & 99.02 & 74.00 & 99.05 & \textbf{74.05} & 98.14 & 72.99 \\
		\midrule
		\multirow{5}{*}{Processed} & RF & 99.79 & 81.89 & 99.82 & 87.28 & 99.90 & \textbf{94.88} \\
		 & KNN & 99.20 & 74.15 & 99.19 & 74.13 & 99.83 & \textbf{82.49} \\
		 & Bernoulli NB & 98.15 & 73.04 & 85.09 & 43.44 & 98.91 & \textbf{73.43} \\
		 & DNN-L (MLP 10--5) & 99.81 & \textbf{82.47} & 99.73 & 74.68 & 98.90 & 73.41 \\
		 & DNN-V (MLP 11--11, SGD) & 99.50 & \textbf{74.36} & 99.02 & 73.66 & 98.85 & 73.37 \\
		\bottomrule
	\end{tabular}

\end{table}

\begin{table}[htbp]
	\centering
	\caption{Run time of feature selection, training (fit on train+validation), and inference (per test record). Medians of three runs for RF, KNN, and Bernoulli NB, single run for DNNs; Apple M3 Max (16 cores).}
	\label{tab:runtime}
	\small
	\begin{tabular}{llrrrrrrr}
		\toprule
		& & \multicolumn{2}{c}{QBSO-FS search} & \multicolumn{3}{c}{Training time (s)} & \multicolumn{2}{c}{Inference ($\mu$s / record)} \\
		\cmidrule(lr){3-4}\cmidrule(lr){5-7}\cmidrule(lr){8-9}
		Data set & Model & Time (s) & Evals & All & TPSO & QBSO & All & QBSO \\
		\midrule
		\multirow{5}{*}{Raw} & RF & 71.2 & 303 & 0.29 & 0.29 & 0.24 & 1.13 & 1.13 \\
		 & KNN & 51.8 & 270 & 0.06 & 0.06 & 0.06 & 10.10 & 15.53 \\
		 & Bernoulli NB & 17.3 & 292 & 0.08 & 0.07 & 0.07 & 0.10 & 0.07 \\
		 & DNN-L (MLP 10--5) & -- & -- & 6.71 & 7.32 & 4.09 & 0.19 & 0.17 \\
		 & DNN-V (MLP 11--11, SGD) & -- & -- & 9.18 & 8.66 & 10.03 & 0.20 & 0.18 \\
		\midrule
		\multirow{5}{*}{Processed} & RF & 96.3 & 551 & 0.21 & 0.22 & 0.17 & 1.29 & 1.31 \\
		 & KNN & 81.5 & 547 & 0.00 & 0.01 & 0.01 & 11.80 & 16.68 \\
		 & Bernoulli NB & 3.1 & 536 & 0.01 & 0.00 & 0.01 & 0.12 & 0.08 \\
		 & DNN-L (MLP 10--5) & -- & -- & 3.72 & 8.57 & 1.45 & 0.18 & 0.17 \\
		 & DNN-V (MLP 11--11, SGD) & -- & -- & 2.21 & 3.34 & 2.25 & 0.18 & 0.17 \\
		\bottomrule
	\end{tabular}
\end{table}

For RF, KNN, and Bernoulli NB, the QBSO-FS subset gives the highest Macro-F1 on both data sets on this split. For Bernoulli NB, however, its margin over all features and over the TPSO subset on the raw data is below 0.2\,pp. The TPSO subset is close to all features for RF and KNN on the raw data, and it is clearly worse for Bernoulli NB on the processed data (85.09\% accuracy versus 98.91\% for the QBSO subset). For the two DNNs, the opposite holds. The TPSO subset performs about as well as all features, whereas the RF-selected QBSO subset lowers accuracy by about 1\,pp. The best overall result, RF with the 7-feature QBSO subset on the deduplicated data (99.90\% accuracy, 94.88\% Macro-F1), exceeds the best of our DNN re-implementations (99.81\% accuracy, 82.47\% Macro-F1) and needs 0.17\,s of training instead of 3.72\,s. The Macro-R of most DNN runs stays below 75\%, which is consistent with the reset-both class being rarely or never predicted.

The QBSO-FS search takes between 3\,s (Bernoulli NB) and 96\,s (RF) and is performed once. It uses 270--551 subset evaluations, each of which is a single fit on the training set. With the reported population size and number of iterations (20 and 20), TPSO evaluates about 400 subsets, each with a 10-fold CV of a Na\"{\i}ve Bayes classifier; its search time was not reported. Once a subset is chosen, the savings in training and inference time are small in absolute terms on this data set. For example, RF training drops from 0.29\,s to 0.24\,s on the raw data. The savings are also not consistent: KNN inference becomes slower with its QBSO subset (10.10 to 15.53\,$\mu$s per record), and MLP training time is governed by the number of iterations until convergence rather than by the input dimension.

%% file: ch4_discussion.tex
\section{Discussion}
\label{sec:discussion}

\paragraph{Effect of leakage on the reported accuracy.}
Although 18\% of the raw test records are exact copies of training records, the test accuracy of RF and KNN is almost the same on the raw and on the deduplicated data (Table~\ref{tab:holdout_results}). The leakage-free accuracies of the selected subsets (99.84\% and 99.90\% for RF, 99.76\% and 99.83\% for KNN) are also close to the biased values of the original experiment (99.87\% and 99.78\%, Table~\ref{tab:bestacc}). This suggests that, for this data set, the optimism in the original \emph{accuracy} figures is small. A plausible reason is that most duplicates belong to the \emph{deny} class (6,945 of 8,362), which is already well separated by port and volume features. However, this comparison does not isolate the two sources of bias. The raw and processed data differ in both duplicates and features, and the original figures come from the first version of this study~\cite{han2021engrxiv}, whose implementation differs from the one used here. Quantifying selection bias and duplicate leakage separately would require re-running the original protocol with the same implementation, and evaluating a deduplicated data set that keeps the original 11 features. Regardless of the size of the effect, the deduplicated held-out protocol is the estimate that should be reported, because it removes both sources of optimism by construction.

\paragraph{Macro-averaged metrics and the reset-both class.}
Accuracy is dominated by the three large classes, whereas the macro-averaged scores are strongly influenced by \emph{reset-both}, which has only 54 records in total and about 11 in each test set. A single misclassified reset-both record changes Macro-R by roughly 2\,pp. This explains the larger and less regular differences in Macro-R and Macro-F1 across settings, for example KNN on the processed data, where precision is near-perfect but Macro-R is 79.40\%. Conclusions about minority-class performance should therefore be drawn with care, and repeated splits or a class-weighted fitness would be needed to optimize them reliably.

\paragraph{Effect of preprocessing on Bernoulli NB.}
Bernoulli NB binarizes every input at zero, so on the raw data most features carry little information (ports, bytes, and packets are almost always non-zero) and accuracy stays at about 80\%. On the processed data, the NAT-source-is-zero and NAT-destination-is-zero flags and the source and destination port ranges are binary or low-cardinality indicators, and accuracy rises to 98.15\% with all features and 98.91\% with the selected subset. Preprocessing is thus far more important for Bernoulli NB than feature selection.

\paragraph{Comparison with previous studies.}
The numbers reported in previous studies (Table~\ref{tab:prior_reported}) should not be compared directly with ours or with each other, because the class definitions, balancing, feature selection data, and averaging differ. The common-protocol re-evaluation (Table~\ref{tab:prior_eval}) therefore gives the more reliable comparison. It also shows that wrapper-selected subsets~\cite{kohavi1997wrappers} are classifier-specific: the RF subset does not transfer to an MLP, and the TPSO subset, chosen with a Na\"{\i}ve Bayes classifier on raw values, does not suit Bernoulli NB on the processed features. A subset selected for one classifier should therefore not be reused for another without re-evaluation.

\paragraph{Run time.}
A motivation of this study was that high-dimensional features take a long time to classify. With only 11 to 16 features, this motivation does not hold for the present data set: the one-off search cost of QBSO-FS (at most about 1.5 minutes) is larger than the training time it saves (Section~\ref{sec:res_prior}). The main benefit of feature selection here is therefore better accuracy and Macro-F1 for RF and KNN, rather than speed. Run time comparisons with previous studies are not possible, because they used different hardware, did not report search times, and in one case reported no times at all.

\paragraph{Limitations.}
Several features are recorded only after the firewall has applied its action. Bytes received and packets received are zero in 99.9\% of deny and 100\% of drop sessions, but in only 9.8\% of allow sessions; elapsed time and the NAT source port are zero in 99.9--100\% of deny and drop sessions, compared with 1.0\% and 1.5\% of allow sessions. The classifiers therefore largely recognize the consequences of the action rather than predict it from information available when the decision is made. This makes the task easier than real-time policy decisions, and limits the direct use of the models to retrospective log analysis and auditing. KNN was run without feature scaling, so part of its gain from feature selection may come from removing features with large numeric ranges, such as ports. The leakage-free experiment uses a single stratified split and a single run of the search, and the QBSO-FS budget of Table~\ref{tab:hyper} is small (3 bees, 3 iterations). Small differences, in particular in the macro-averaged scores, are within the variability that can be expected from a different split or run. In addition, apart from the TPSO subset of \citet{verma2024tpso}, which we evaluated but did not re-run, QBSO-FS is compared only with using all features. Comparisons with simpler feature selection baselines, such as ranking by random forest importance or mutual information, or random subsets evaluated with the same budget, are not included. Repeating the protocol over several splits and runs, testing whether the improvement is significant, and adding such baselines are left for future work.

%% file: ch5_conclusion.tex
\section{Conclusion}
In this paper, we applied QBSO-FS to select features for firewall log classification. The first version of this study~\cite{han2021engrxiv} reported that the selected features outperformed all features. However, those results were obtained by selecting and evaluating feature subsets on the same CV folds of a data set that contains 12.8\% duplicated records, and they are therefore optimistically biased.

In the extended experiment, we removed exactly duplicated records and selected features on a separate validation set, keeping the test set untouched. Under this leakage-free protocol, QBSO-FS with the original hyper-parameters still selected compact subsets (5--7 features). On a single split, these subsets clearly improved Macro-F1 over all features for KNN on both data sets and for RF on the deduplicated data. For RF on the raw data and for Bernoulli NB, the differences were small and may not be robust, so in these cases the subsets should be regarded as matching, rather than exceeding, the full feature set. RF with 7 selected features on the deduplicated data reached 99.90\% accuracy and 94.88\% Macro-F1. These results suggest that, at least for KNN and RF, the benefit of QBSO-FS feature selection for firewall log classification is not merely an artifact of test-set leakage. Confirming the size of the benefit over multiple splits and against other feature selection baselines is left for future work.